\documentclass[11pt]{article}

\usepackage[preprint]{acl}

\usepackage{times}
\usepackage{latexsym}
\usepackage[T1]{fontenc}
\usepackage[utf8]{inputenc}
\usepackage{microtype}
\usepackage{inconsolata}
\usepackage{graphicx}

\usepackage{wrapfig}
\usepackage{booktabs}
\usepackage{tikz}
\usepackage{amssymb}
\usetikzlibrary{arrows.meta}
\usepackage{listings}
\usepackage{placeins}
\usepackage{float}
\usepackage{tabularx}
\usepackage{array}
\usepackage{amsmath}
\usepackage{todonotes}
\usepackage{pifont}

\newcommand{\cmark}{\ding{51}}
\newcommand{\xmark}{\ding{55}}
\newcommand{\absmark}{--}

\newcommand{\best}[1]{\textbf{#1}}
\newcommand{\runner}[1]{\underline{\smash{#1}}}

\newcommand{\inlinenote}[4][]{%
  \todo[
    inline,
    author=#2,
    color=#3,
    size=\scriptsize,
    caption={},
    #1
  ]{#4}%
}

\renewcommand{\inlinenote}[4][]{}

\newcommand{\reut}[2][]{\inlinenote[#1]{Reut}{violet!40}{#2}}
\newcommand{\amit}[2][]{\inlinenote[#1]{Amit}{red!40}{#2}}

\title{From Abstraction to Action:\\ 
DSLs for Natural Language Abstraction Interpretation and Execution}

\author{
  Amit Lisha \and Reut Tsarfaty \\
  Bar-Ilan University, Israel \\
  \texttt{\{amit.lisha,reut.tsarfaty\}@biu.ac.il}
}

\begin{document}
\maketitle

\begin{abstract}
Natural-language instructions often rely on high-level abstractions, such as asking an agent to ``stack  blocks in a pyramid.''
Interpreting abstract instructions requires models to both identify 
the abstraction  intended by the speaker, and unroll it into the many concrete actions that instantiate it.
Language models struggle here: prior work has shown that performance degrades as the level of abstraction in the instruction rises.
To bridge the gap between natural language abstraction and concrete actions, we propose to borrow from coding environments the notion of
a {\em domain-specific language} (DSL), an explicit representation layer whose documented primitives name recurring abstractions 
that expand into concrete actions. 
We demonstrate the efficacy of DSLs for NL interpretation on \textsc{Hexagons}, a visually-grounded instruction-following task graded
for abstraction.
We further ask whether language models can induce such DSLs themselves from domain data.
Our results show that while code generation consistently improves over direct inference, code generation over a human-curated DSL yields dramatic gains, up to 10 absolute F1 points, reaching the human skyline, with gains concentrated at {\em higher} abstraction levels. We confirm a similar pattern on the \textsc{Text2CAD} benchmark. At the same time, model-induced DSLs recover only part of this gain, remaining inconsistent across runs and short of their human-curated counterpart.
Overall, the right representation  closes much of the gap caused by high abstraction --- but inducing that layer of representation automatically is where models still fall short, and where we see a strong opportunity for the community to make progress.\footnote{We open-source our data, code, DSLs for future research.}

\end{abstract}

\section{Introduction}
\begin{figure*}[t]
\makebox[\textwidth][l]{%
\resizebox{\textwidth}{!}{%
\begin{tikzpicture}[
    instrbox/.style={
        draw=blue!25,
        rounded corners=3pt,
        fill=blue!5,
        text=black,
        align=center,
        inner sep=9pt,
        font=\small
    },
    codebox/.style={
        draw=black!18,
        rounded corners=3pt,
        fill=gray!4,
        align=left,
        font=\tiny\ttfamily,
        inner sep=6pt
    },
    smallbox/.style={
        draw=black!18,
        rounded corners=3pt,
        fill=#1,
        minimum width=1.7cm,
        minimum height=0.58cm,
        align=center,
        font=\scriptsize\bfseries
    },
    sec/.style={
        font=\normalsize\bfseries,
        text=black!75
    },
    arr/.style={
        -{Stealth[length=3.5pt]},
        semithick,
        draw=black!40
    },
    darr/.style={
        -{Stealth[length=3.5pt]},
        semithick,
        densely dashed,
        draw=black!25
    },
    arrl/.style={
        font=\tiny\itshape,
        text=black!40
    },
    tinybox/.style={
        draw=black!18,
        rounded corners=3pt,
        fill=#1,
        minimum width=1.45cm,
        minimum height=0.46cm,
        align=center,
        font=\tiny\bfseries
    },
]

\def\pL{4.7}
\def\pR{15.9}

\def\yA{0.95}
\def\yB{-2.35}
\def\yTopShift{0.4}   
\def\yHexShift{1.5}   
\def\yCShift{2.6}

\def\xCShift{0.6}   

\def\greenBottom{-2.05}
\def\yCBoxesShift{0.18}   

\pgfmathsetmacro{\yADisp}{\yA+\yTopShift}
\pgfmathsetmacro{\yBDisp}{\yB+\yHexShift}
\pgfmathsetmacro{\yFork}{(\yADisp+\yBDisp)/2}

\pgfmathsetmacro{\cTop}{-5.15+\yCShift}


\node[instrbox, text width=3.35cm] (instr) at (1.95, \yFork) {%
    \textit{``Draw a \textbf{red flower} with a green center,
    centered at the fifth row and seventh column.
    Then place a \textbf{color inverted} one directly
    \textbf{below} it.''}%
};

\coordinate (spine)  at (4.45, \yFork);
\coordinate (spineA) at (4.45, \yADisp);
\coordinate (spineB) at (4.45, \yBDisp);

\draw[arr] (instr.east) -- (spine);
\draw[semithick, draw=black!40] (spineA) -- (spineB);


\fill[red!2, rounded corners=5pt, draw=black!8]
    (\pL, 2.45) rectangle (\pR, 0.3);

\def\titlePad{0.1}

\node[sec, anchor=west] at (\pL+\titlePad, 2.10) {(a) Direct Prediction};

\node[smallbox={orange!12}] (llmA) at (5.9, \yADisp) {LLM};
\draw[arr] (spineA) -- (llmA.west);

\node[inner sep=0pt] (imgA) at (14.40, \yADisp) {%
    \includegraphics[width=2.00cm]{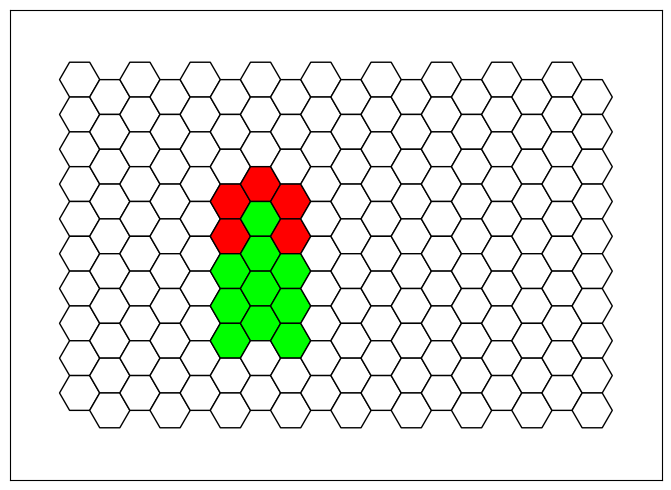}%
};

\draw[arr] (llmA.east) -- (imgA.west)
    node[arrl, midway, above=0pt] {predicts tiles};

\node[font=\scriptsize\bfseries, text=red!60!black]
    at ([yshift=-0.12cm]imgA.south) {$\times$ Incorrect};

\fill[green!2, rounded corners=5pt, draw=black!8]
    (\pL, 0.15) rectangle (\pR, -2.05);

\def\titlePad{0.1}

\node[sec, anchor=west] at (\pL+\titlePad, -0.2) {(b) \textsc{HexaGen}};

\node[smallbox={orange!12}] (llmB) at (5.90, \yBDisp) {LLM};
\draw[arr] (spineB) -- (llmB.west);

\node[tinybox={blue!10}] (dslB) at (5.90, {\yBDisp-0.8}) {DSL};
\draw[arr] (dslB.north) -- (llmB.south);

\node[codebox, text width=4.55cm, anchor=west] (code) at (7.65, {-2.35+\yHexShift}) {%
with Game() as g:\\
\;\;c = Tile(5,7).draw('green')\\
\;\;flower = Circle(c,r=1).draw('red') + c\\
\;\;flower.copy\_paste('down').recolor\\
\;\;\;\;\;\;\;\;\;\;\;\;\;\;\;\;\;\;\;\;\;(\{'red':'green','green':'red'\})%
};

\draw[arr] (llmB.east) -- (code.west)
    node[arrl, pos=.50, above=0pt] {generates};

\node[inner sep=0pt] (imgB) at (14.40, {-2.35+\yHexShift}) {%
    \includegraphics[width=2.05cm]{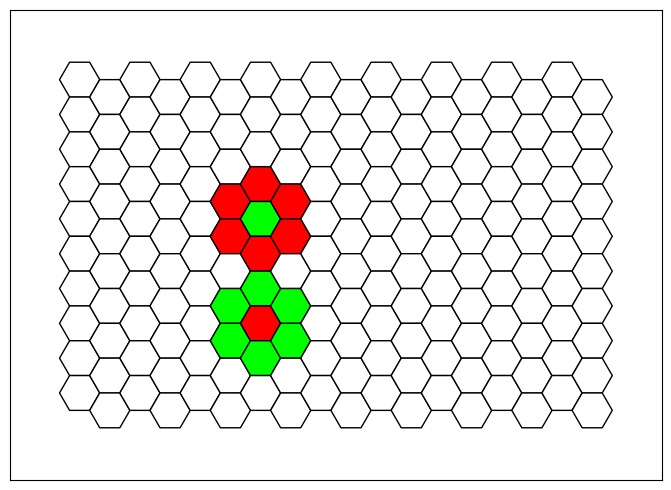}%
};
\draw[arr] (code.east) -- (imgB.west)
    node[arrl, pos=.50, above=0pt] {executes};

\node[font=\scriptsize\bfseries, text=green!50!black]
    at (14.35, {-3.24+\yHexShift}) {$\checkmark$ Correct};


\end{tikzpicture}%
}%
}
\caption{%
Abstract instructions compress many low-level drawing actions on the board.
\textbf{(a)}~Direct prediction must recover this structure directly, in an end-2-end fashion.
\textbf{(b)}~\textsc{HexaGen} makes task-relevant abstractions explicit, letting the model generate code that executes to the correct output.
}
\label{fig:intro}
\end{figure*}

Language models are increasingly deployed as agents that must \emph{act} on natural-language instructions: editing, drawing, operating software, manipulating an environment, and a lot more. However, 
the instructions users write are rarely explicit sequences of actions. A user asking an agent to ``add a `total' row to each table in the report'' expects it to recover the many concrete edits compressed into this phrase, rather than be instructed for each one.

This tendency for {\em compression} is unsurprising: human speakers typically provide the minimal amount of information that is needed for a successful interpretation \citep{grice1975logic}. Speakers thus convey multiple actions through {\em abstraction}, defined by \citet{Wing2010ComputationalTW}  as ``the representation of many instances through a single higher-level object, capturing essential structure while suppressing lower-level details.'' 

For a language model to carry out abstract instructions, it must identify the abstract concepts and reconstruct the sequence of  actions they leave implicit \citep{walk-the-talk, artzi-zettlemoyer-2013-weakly}. Models are unevenly good at this: performance declines as instructions grow more abstract \citep{ALFRED20, ruis2020gscan, wu-etal-2021-reascan, saycan2022arxiv}, particularly evident in \textsc{Hexagons} \citep{lachmy-etal-2022-draw}, where instructions are annotated with abstraction levels.

This suggests a simple hypothesis: if the gap between language and action arises because abstract instructions leave task-relevant steps implicit, then unfolding those abstractions explicitly should help bridge it.
Code generation is a natural first step to explore \citep{chen2021codex, codeaspolicies2022}: programming languages provide built-in mechanisms for unfolding abstraction --- variables, loops, functions, conditions --- so mapping instructions to code gives the model computational structure that exposes the structures implicit in language. 

However, general-purpose code may not suffice: the model may still need to recover \emph{task-specific} abstractions from scratch, e.g., what counts as a ``table'' or a ``header'' in a document, and reflect them as concrete operations.

In business environments, such recurring domain concepts are often provided as reusable operations through a {\em domain-specific language} (DSL) \citep{mernik2005} that is then used in code to map abstraction to action.
%
In this work we ask: (i) can language models use task-specific DSLs  to  effectively follow {natural language instructions}? and, moreover, (ii)~may language models come up with such DSLs themselves?

\begin{figure}[t]
  \centering
\scalebox{0.9}{
  \setlength{\fboxsep}{3pt}
  \fbox{%
    \begin{minipage}{\dimexpr\columnwidth-2\fboxsep-2\fboxrule\relax}
    \footnotesize
    \setlength{\parindent}{0pt}
    \begin{minipage}[c]{0.50\linewidth}
      \textbf{1.} Make a blue flower of 6 tiles surrounding the 5th tile in the 10th column.
    \end{minipage}%
    \hfill
    \begin{minipage}[c]{0.46\linewidth}
      \centering
      \includegraphics[width=\linewidth,height=2.3cm,keepaspectratio]{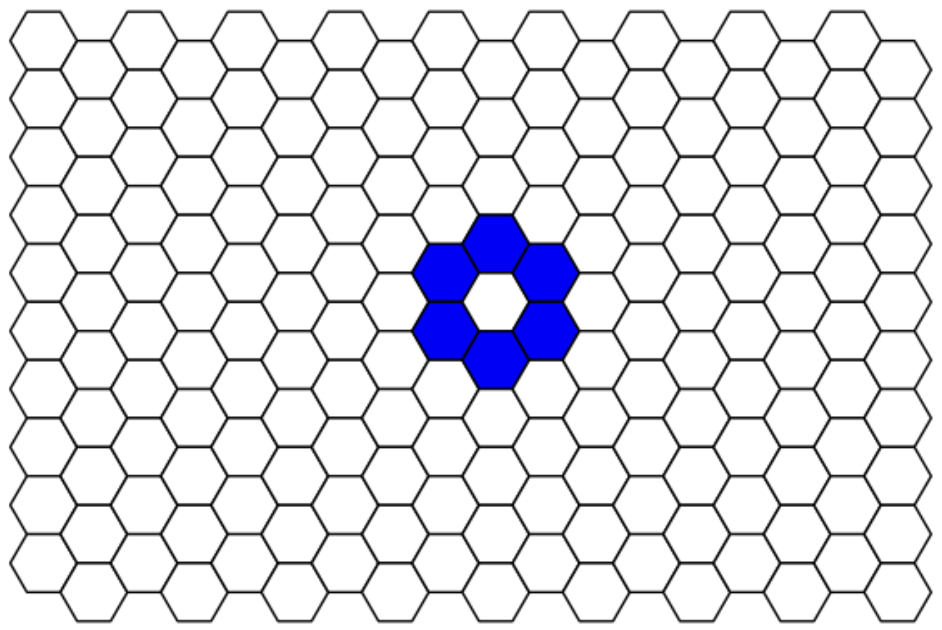}
    \end{minipage}
    \par\vspace{0.4em}
    \begin{minipage}[c]{0.50\linewidth}
      \textbf{2.} Make 4 orange flowers connecting to the original, on each of the outermost corners.
    \end{minipage}%
    \hfill
    \begin{minipage}[c]{0.46\linewidth}
      \centering
      \includegraphics[width=\linewidth,height=2.3cm,keepaspectratio]{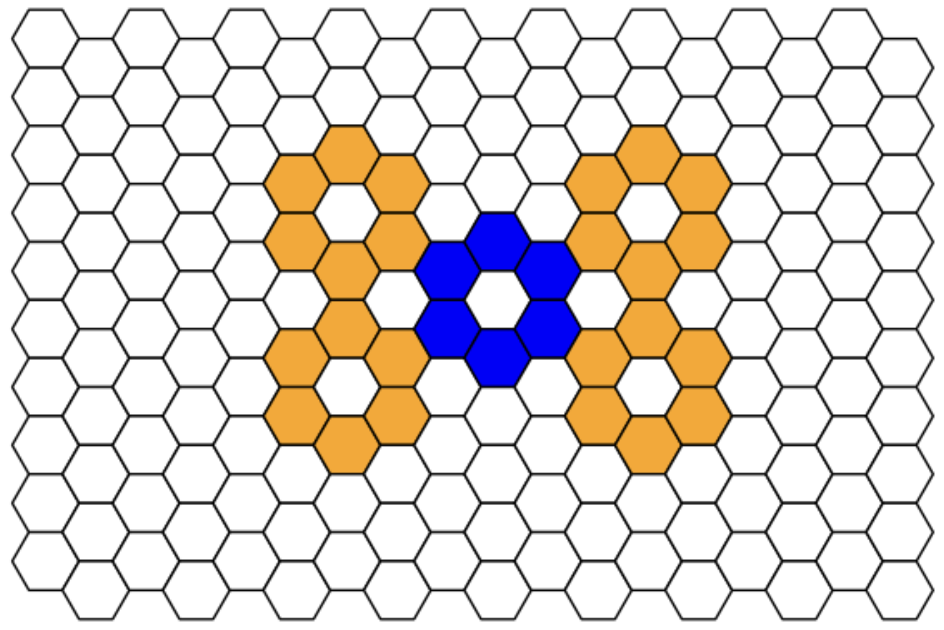}
    \end{minipage}
    \end{minipage}%
  }}
  \caption{The {\sc Hexagons} task illustration: Natural language instructions and respective target board states.}
  \label{fig:task_example}
\end{figure}

Our primary empirical investigation focuses on \textsc{Hexagons}, a visually grounded instruction-following benchmark designed to elicit abstraction in natural language. Its instructions refer to higher-level structures: complex objects, patterns, arrangements; and successful execution depends on producing the right low-level  actions (Figure~\ref{fig:task_example}). Furthermore, \textsc{Hexagons}'s abstraction-level annotation lets us control performance on abstraction, and isolate where the interpretation  breaks down.

We investigate the first question via a three-way comparison of strategies to solve the {\sc Hexagons} board: direct inference, unrestricted Python generation, and Python generation over a DSL. In particular, we assess performance given \textsc{HexaGen}, a human-authored Python DSL we designed for this purpose, which defines decomposable abstractions on top of 
the {\sc Hexagon} tiles: named shapes, spatial queries and geometric transformations.
The model may write an ordinary Python program and invoke DSL functions as needed, retaining full interpretive responsibility while offloading low-level domain-specific reasoning to the DSL (Figure~\ref{fig:intro}).

Our  results show that a human-curated DSL indeed leads to substantial performance gains, closing much of the gap with the human skyline, an effect we show on both {\sc Hexagon} (Section~\ref{sec:results-execution})
and a {\sc Text2CAD}  real-world scenario (Section~\ref{sec:text2cad}).
This success  
begs our second question, namely,  can LLMs  induce such DSLs themselves?

We investigate this by augmenting the {\sc Hexagons} task training data, instructions paired with the denotations on the board, with a minimal API covering only low-level drawing actions. Models are asked to identify recurring patterns and implement respective DSL functions, producing executable Python code and documentation for downstream use. Replacing \textsc{HexaGen} with such induced DSLs shows mixed results relative to a minimal-API baseline, with improvements varying across models, but consistently underperforming the {\em human-curated DSL} counterpart.

Taken together, our experiments show that interpreting and executing natural language instruction using a  curated DSL consistently outperforms both direct inference and unrestricted code generation, with the largest gains where interpretation is hardest: at higher abstraction levels.
At the same time, we demonstrate that current LLMs fall short of human experts on the task of reliably inducing such abstraction layers themselves from natural language domain-specific data --- leaving the automatic discovery of task-relevant abstraction layers from natural language texts as an open challenge for further investigation.

\section{From Abstraction to Action:
}\subsection{
Challenge and  Key Hypotheses}

Interpreting and grounding abstract instructions poses a genuine challenge for LLMs, as it requires mapping natural-language abstractions to their underlying implicit and concrete actions to execute. Here we argue that tackling this challenge is inherently multifaceted.

First we observe that formal coding languages are a natural, unambiguous way to express abstractions and how they unfold. Providing the model access to explicit computation --- variables, loops, functions --- alongside the language utterances,  may help recover the actions that natural language leaves implicit. We therefore hypothesize that mapping instructions to executable code (rather than, e.g., relying on thinking trajectories) is a better strategy than direct inference on the task at hand.

Following this intuition, we cast instruction-following as a {\em program-synthesis} problem over a grounding environment: given a natural-language instruction~$x$ and an initial  state~$s_0$, the model must produce a program~$p$ whose execution yields the target state~$s^*$. 

Yet, code generation in a general-purpose language does not fully close the abstraction gap for a specific task.
Plain Python lacks task-specific  notions, (in {\sc Hexagons} (\textsection\ref{hex}), for example, these may be hexagonal grids, spatial neighbors, or geometric objects,) forcing the model to learn and implement every domain concept from scratch. 
Generally, 
we argue that code generation alone may not suffice for many real-world tasks, whose relevant abstractions are often domain-specific and lack, or are divorced from, their interpretation in a general-purpose language. So, just as SQL encapsulates relational operations like joins and aggregations rather than requiring repetitive iterations, a {\em domain-specific language} (DSL) can encapsulate recurring domain concepts as reusable operations \citep{mernik2005}. 

We draw from this a key conclusion: 
while code generation provides the machinery of computation, what is still missing is a way to encapsulate recurring domain concepts as reusable operations that the model can compose and reuse rather than re-derive. For this, a DSL can help.

\subsection{The \textsc{Hexagons} benchmark}
\label{hex}

The \textsc{Hexagons} benchmark \citep{lachmy-etal-2022-draw} is an instruction-following benchmark designed to elicit abstraction in natural language, derived from a collaborative drawing game in which one player describes how to recreate an image on a hexagonal grid while the other follows the instructions on an empty board. Since players are rewarded for both accuracy and speed, speakers naturally resort to compressed, abstract descriptions characteristic of natural communication --- referring to objects, loops, spatial transformations, and conditions --- each standing for many low-level actions, with missing details left to be filled by the hearer.

The core challenge in solving the {\sc Hexagons} task is that carrying out such instructions requires more than understanding explicit content: the model must \emph{identify} the abstract constructions in the instruction, \emph{ground} them in exact tile coordinates, and \emph{execute} the implied actions correctly. As illustrated in Figure~\ref{fig:task_example}, ``a blue flower'' compresses a spatial arrangement of seven tiles into a single reference, while ``on each of the outermost corners'' implicitly specifies four locations dependent on the geometry of the previously constructed object. These stages are tightly coupled --- errors in interpretation propagate through grounding into execution --- and each grows harder as abstraction increases, as confirmed empirically by \citet{lachmy-etal-2022-draw}. \textsc{Hexagons} therefore provides a controlled testbed for studying how to improve the \emph{interpretation}, \emph{grounding}, and \emph{execution} of abstract instructions in contemporary LLMs.

\section{DSLs for Natural Language Abstraction}
\label{sec:dsl}

\subsection{From General Code to Task-Specific Libraries}
The notion of a {\em domain-specific language} (DSL) is known from   coding environments, and it refers to a specialized library tailored to a particular application domain, for defining, breaking down and spelling out domain-specific abstract notions \citep{mernik2005}.  
By raising the level of abstraction to match the problem at hand, DSLs reduce both the effort and the error rate of programming within their target domain.  
In our setting, we aim to use DSLs for interpreting naturally occurring  language instructions in different domains, likewise reducing  the effort and  errors in interpreting and following   natural language instructions.

On the way to obtain this, we face  several concerns in the design of DSLs for NL. The first is \emph{representation}: the DSL should provide first-class constructs for the abstractions that genuinely organize solutions in the target domain. The second is \emph{transparency}: because our DSL is to be consumed by a language model, what each primitive does and how it should be called must be clearly evident from its name, signature, and documentation \citep{quexploration}. The third is \emph{robustness}: constrained semantics, safe defaults, and failure-resistant interfaces should ensure that when a program fails, it fails because the intended construction is wrong, not because the programming interface is fragile.

\paragraph{From Hexagons to HexaGen}

We apply these principles to \textsc{Hexagons} to deliver \textsc{HexaGen}, an expert-curated Python DSL that supplies the missing abstraction layer for following instrucions on the \textsc{Hexagons} board. On the \emph{representational} side, we identified its core primitives by manually analyzing training examples, guided by the abstraction categories in \citet{lachmy-etal-2022-draw}. For \emph{transparency}, the API exposes non-overlapping, self-explanatory methods with thorough documentation. For \emph{robustness}, it validates inputs eagerly, provides safe defaults, and raises descriptive exceptions for common pitfalls, thus errors reflect flawed usage, not interface fragility.
The full DSL specification, including all constructors, actions, queries, and their signatures, is provided in 
in Appendix~\ref{app:hexagen_api}. A representative example is given in Figure~\ref{fig:intro}.

\subsection{Inducing DSLs from Data}
\label{sec:dsl_induction}

Creating useful and adeqaute DSLs, such as \textsc{HexaGen}, is a human-intensive and costly engineering effort. The theoretical benefits of DSLs in general, and of  \textsc{HexaGen} in particular, beg a natural follow-up question: can LLMs induce such  abstraction layers themselves, given only examples of the task?

We formulate this problem as follows. Let $\mathcal{A}_0$ denote a \emph{foundation library} exposing only the task-specific primitives. 
Let $\mathcal{D} = \{(x_i, s_i)\}_{i=1}^{N}$ be a training set of paired natural-language instructions and their denotation (in {\sc Heaxgons} (\textsection\ref{hex}) for example, these are instruction paired with target board states represented as  tile-level descriptions). 
The induction task is, given  $\mathcal{A}_0$ and $\mathcal{D}$ produce an extended library $\mathcal{A} \supseteq \mathcal{A}_0$, consisting of new callable functions, together with API documentation $\mathtt{Doc}(\mathcal{A})$ suitable for downstream prompting.

We suggest that the induction pipeline be comprised of three stages: {\bf(i) abstraction discovery}, which identifies recurring, non-trivial computational patterns in the instruction corpus, {\bf (ii) method implementation}, which translates the discovered method specifications into validated executable code on top of the foundation library, and {\bf (iii) documentation authoring}, which produces a compact API specification that serves as the sole downstream prompting interface. 
\reut{commented stuff here - move to exp settings}

\begin{table*}[t]
\centering
\setlength{\tabcolsep}{4pt}
\resizebox{\textwidth}{!}{%
\begin{tabular}{l c c c c c c c c c c c c}
\toprule
& \multicolumn{8}{c}{Proprietary Models}
& \multicolumn{4}{c}{Open-Source Models} \\
\cmidrule(lr){2-9}
\cmidrule(lr){10-13}
& \multicolumn{2}{c}{GPT-5.1 (Low)}
& \multicolumn{2}{c}{GPT-5.1 (High)}
& \multicolumn{2}{c}{Gemini 3 Pro (Low)}
& \multicolumn{2}{c}{Gemini 3 Pro (High)}
& \multicolumn{2}{c}{Qwen3 30B}
& \multicolumn{2}{c}{Gemma3 27B} \\
\cmidrule(lr){2-3}
\cmidrule(lr){4-5}
\cmidrule(lr){6-7}
\cmidrule(lr){8-9}
\cmidrule(lr){10-11}
\cmidrule(lr){12-13}
& F1 & EM & F1 & EM & F1 & EM & F1 & EM & F1 & EM & F1 & EM \\
\midrule
Tiles
& 0.683{\scriptsize$\pm$0.006} & \runner{0.236{\scriptsize$\pm$0.018}}
& \runner{0.765{\scriptsize$\pm$0.016}} & \runner{0.322{\scriptsize$\pm$0.032}}
& 0.743{\scriptsize$\pm$0.009} & 0.301{\scriptsize$\pm$0.009}
& 0.745{\scriptsize$\pm$0.02} & 0.382{\scriptsize$\pm$0.037}
& 0.556{\scriptsize$\pm$0.016} & \runner{0.177{\scriptsize$\pm$0.013}}
& 0.407{\scriptsize$\pm$0.005} & 0.123{\scriptsize$\pm$0.007} \\
Unrestricted Python
& \runner{0.708{\scriptsize$\pm$0.011}} & 0.231{\scriptsize$\pm$0.009}
& 0.757{\scriptsize$\pm$0.009} & 0.296{\scriptsize$\pm$0.009}
& \runner{0.807{\scriptsize$\pm$0.003}} & \runner{0.376{\scriptsize$\pm$0.040}}
& \runner{0.793{\scriptsize$\pm$0.021}} & \runner{0.419{\scriptsize$\pm$0.000}}
& \runner{0.587{\scriptsize$\pm$0.012}} & 0.172{\scriptsize$\pm$0.007}
& \runner{0.495{\scriptsize$\pm$0.005}} & \best{0.145{\scriptsize$\pm$0.000}} \\
Human-Curated DSL
& \best{0.774{\scriptsize$\pm$0.021}} & \best{0.371{\scriptsize$\pm$0.000}}
& \best{0.831{\scriptsize$\pm$0.010}} & \best{0.441{\scriptsize$\pm$0.025}}
& \best{0.825{\scriptsize$\pm$0.002}} & \best{0.457{\scriptsize$\pm$0.024}}
& \best{0.853{\scriptsize$\pm$0.016}} & \best{0.511{\scriptsize$\pm$0.046}}
& \best{0.663{\scriptsize$\pm$0.019}} & \best{0.237{\scriptsize$\pm$0.033}}
& \best{0.512{\scriptsize$\pm$0.007}} & \runner{0.140{\scriptsize$\pm$0.007}} \\
\bottomrule
\end{tabular}%
}
\caption{Hexagons execution results across all models and strategies. We report F1 and board-level exact match (EM) as mean $\pm$ standard deviation over three runs.}

\label{tab:main-results}
\end{table*}

\section{Experimental Setup}
\label{sec:experimental_setup}


\noindent\textbf{Goal.}
We set out to empirically investigate two questions: whether a task-aligned DSLs improve LLM performance over plain code generation, and whether LLMs can induce useful DSLs themselves.

\noindent\textbf{Data.}
We use the original \textsc{Hexagons} dataset \cite{lachmy-etal-2022-draw}, consisting of 620 procedures and 4,177 language instructions overall, where each procedure is a sequence of instructions paired with the corresponding board state. Later, we also use the Text2CAD \cite{text2cad} benchmark.

\noindent\textbf{Task.}
The \textsc{Hexagons} task is defined as follows. The input is a procedure describing a complex image, composed of multiple instructions. The procedure is processed all at once.\footnote{Appendix~\ref{app:full_procedure_benchmark} compares this setup with the original step-by-step benchmark protocol of \cite{lachmy-etal-2022-draw}.}
We evaluated models under one of three execution scenarios:
\textbf{(i)~Direct Prediction}, which outputs a list of \texttt{(row, col, color)} tuples specifying the tiles to paint; \textbf{(ii)~Unrestricted Python Generation}, which outputs Python code implementing \texttt{get\_tiles()} to return the same tuples; and finally, \textbf{(iii)~DSL-based Code Generation}, where the input includes the DSL API documentation and the output is  code using the DSL, whose execution yields the target. 

\noindent\textbf{Prompts.}
All scenarios use a two-phase prompt: a system prompt defining the board, colors, output requirements, and scenario-specific information, if any (API specification, tuple format, or board dimensions), and a user message containing the instructions. The exact 
are stated in Appendix \ref{app:experiment_prompts}.

\noindent\textbf{Models.}
\label{sec:models}
We evaluate both proprietary and open-weight  models,  both \texttt{low} and \texttt{high} reasoning (thinking) settings: \textbf{GPT-5.1} and \textbf{Gemini-3-Pro} (high/low), \textbf{Gemini-3.1-Pro} (high), \textbf{Qwen3-30B-A3B}~(a reasoning model) \citep{qwen3}, and \textbf{Gemma-3-27B-IT}~\citep{gemma_2025}. The open  models are served locally via vLLM on 2 A100 GPUs.

\noindent\textbf{Inference.}
\label{sec:implementation}
All models are decoded with temperature $0$ and a maximum output length of 15K tokens. Generated programs are executed in an isolated environment with a 60-sec.\ timeout and up to 3 retries conditioned on the resulting exception message.

\noindent\textbf{Metrics.}
Following \citet{lachmy-etal-2022-draw}, we report F1 and exact match on the final board state. In the all-at-once setting, board-level F1 equals action-level F1 in \citet{lachmy-etal-2022-draw}. Table~\ref{tab:main-results} reports mean $\pm$ standard deviation over three execution runs. Table~\ref{tab:dsl-induction} reports mean $\pm$ standard deviation across three independently induced libraries, each evaluated over three execution runs.

\section{Does Explicit Abstraction Help?}
\label{sec:results-execution}

\paragraph{Results}
Table~\ref{tab:main-results} presents execution results across all models and strategies. Overall in terms of F1, \emph{Unrestricted Python Generation} improves over \emph{Direct Prediction} in five of six settings, while exact match remains broadly similar overall. This suggests that generating an executable program is often more effective than predicting tiles directly. 
Then, \emph{DSL-based Code Generation} pushes the results further, essentially closing the gap with the human skyline of F1=82.06 reported by \citet{lachmy-etal-2022-draw}. It achieves the highest F1 across all six models and the highest exact match in five of six, indicating that explicit, executable abstractions improve grounding beyond code execution alone.
We further see that weaker models benefit more from the DSL-based strategy: relative to \emph{Direct Prediction}, the three weakest models improve F1 by 13.3--25.8\%, compared with 8.6--14.5\% for the three strongest.

\begin{figure}[t]
  \centering
  \includegraphics[width=\linewidth]{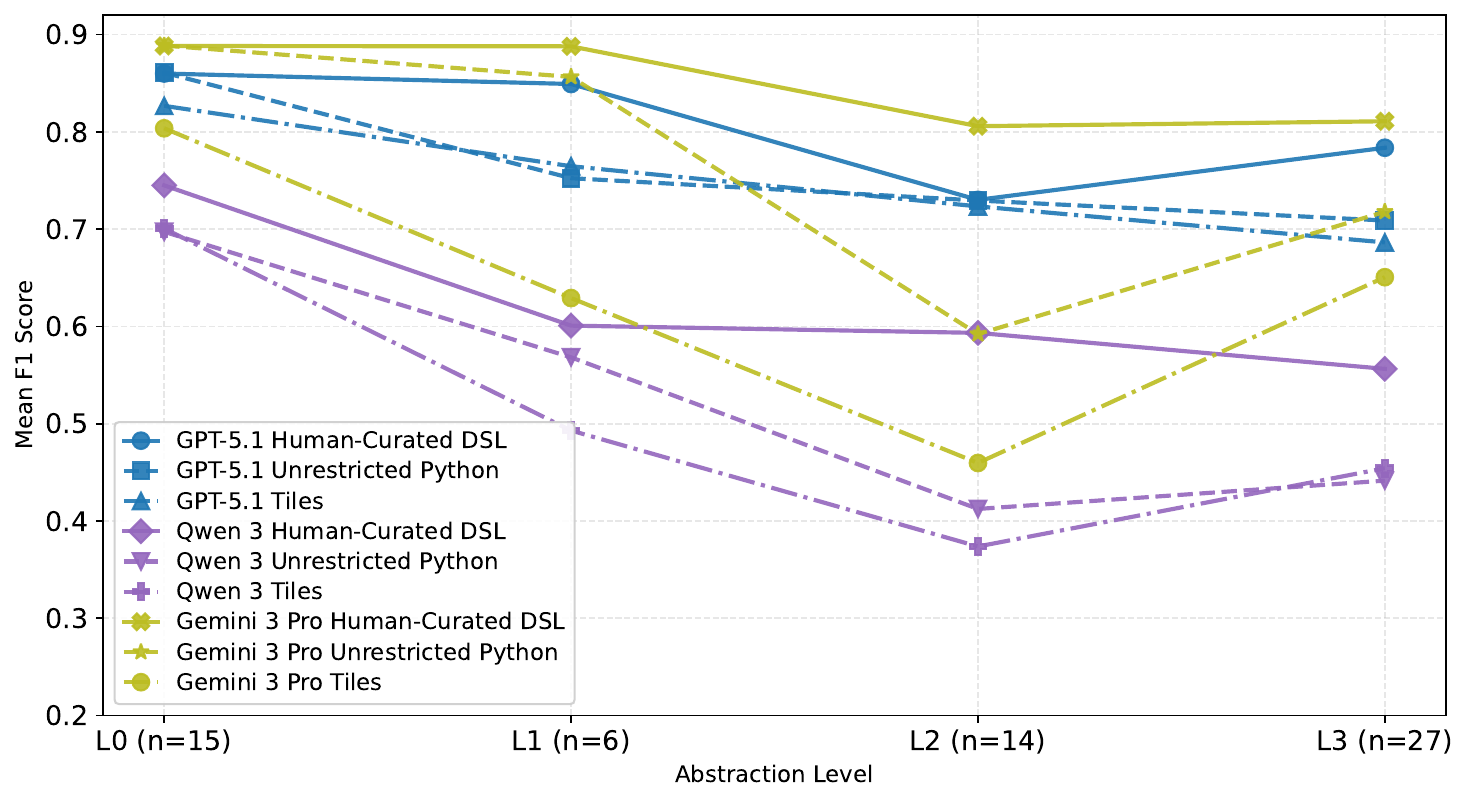}
  \caption{F1 as a function of abstraction level under the 3 strategies.}
  \label{fig:abstraction-degradation}
\end{figure}

\paragraph{Analysis}
\citet{lachmy-etal-2022-draw} showed that model performance degrades as abstraction increases. We now ask the respective question, whether the DSL is especially helpful in the  harder, more abstract, scenarios. Figure~\ref{fig:abstraction-degradation} suggests that it is indeed the case. At Level~0 the DSL method performs on par with the other strategies for every model, but at Levels~2--3 it leads by up to 0.21 F1, essentially showing that the more abstract the instruction, the more the DSL helps.

\begin{figure*}[t]
  \centering
  \includegraphics[width=0.95\textwidth]{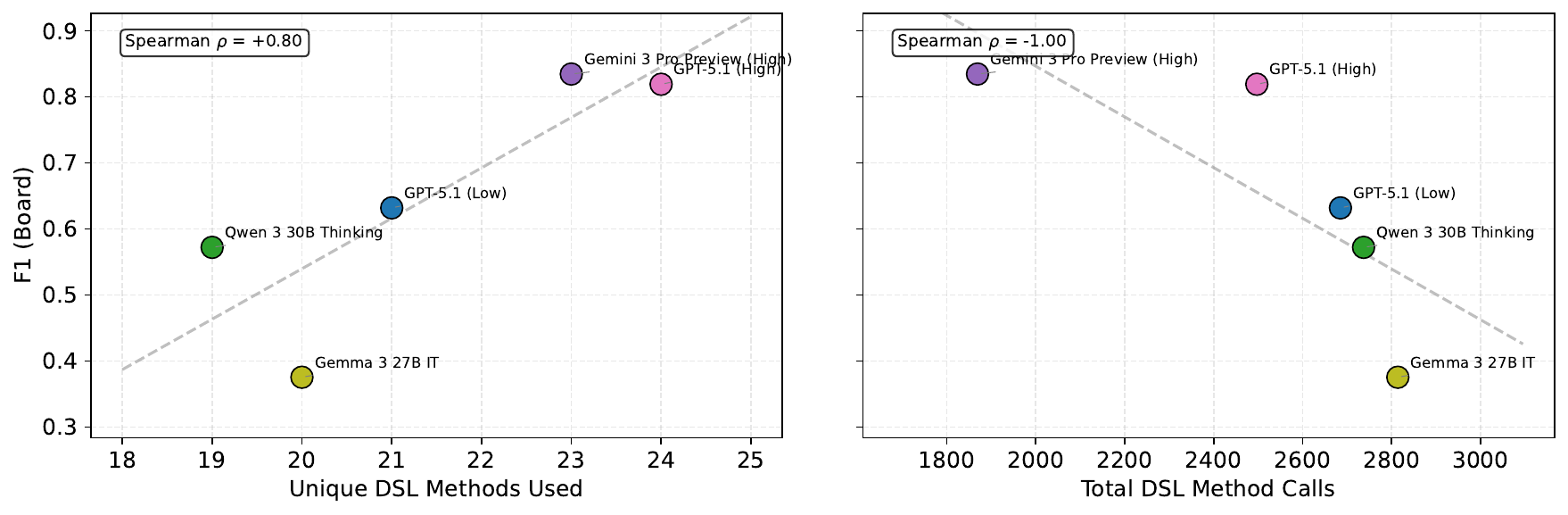}
  \caption{DSL usage patterns vs.\ performance, for 5 model configurations sorted by overall F1. \textbf{Left:} breadth of DSL usage (unique methods invoked) against board F1. \textbf{Right:} volume of DSL usage (total method calls) against board F1.}
  \label{fig:dsl-usage}
\end{figure*}

We next ask \emph{how} models use the DSL.
Figure~\ref{fig:dsl-usage} reports
the \emph{breadth} of DSL usage (i.e., the number of {\em unique} DSL methods invoked, out of the 41 available) and the \emph{volume} of DSL usage (i.e., the total number of DSL method calls issued), each plotted against board F1.
In terms of {\em breadth}, stronger models use more of the DSL; Gemini~3~Pro (High) and GPT-5.1 (High) invoke 23--24 unique methods, versus 19--20 for Gemma~3 and Qwen~3 (Spearman $\rho=+0.80$ with F1). {\em Volume} runs the other way, and tracks F1 almost perfectly (Spearman $\rho=-1.00$): weaker models issue substantially more  DSL calls. 
This suggests that weaker models rely on fewer methods and call them more repetitively, while stronger models spread their calls across a wider slice of the DSL and invoke it more sparingly overall, hinting at more targeted abstraction selection. 
All in all, the DSL helps all models, but the gain scales with the model's ability to exploit the structure it provides.

Finally, we test whether the models' errors stem from lacking visual feedback. Showing the board after each instruction does not help: when adding the visual signal, for \textsc{HexaGen}-based generation the difference is negligible, for direct tile prediction performance degrades markedly (Appendix~\ref{app:vision}).

\section{Does This Hold Beyond \textsc{Hexagons}?}
\label{sec:text2cad}
To test whether our results generalize beyond the {\sc Hexagons} task, we turn to Text2CAD \citep{text2cad} with its CadQuery annotations \citep{text2cadquery}, where each object is paired with design instructions at four levels of abstraction, from \emph{expert} to \emph{abstract}. Taking CadQuery as the foundation library, we hand-curate a small DSL of eight recurring abstractions following the design principles of Section~\ref{sec:dsl}. We follow the evaluation setup of \citet{text2cad} and simplify by using a single protocol across all four abstraction levels: a VLM judge compares renders of the DSL and direct programs pairwise, with position randomized, and returns a preference or declares the pair indistinguishable. Programs that fail to execute count as losses, and pairs where both fail are dropped. Full details are provided in Appendix~\ref{app:text2cad}.
\begin{table}[t]
\centering
\footnotesize
\setlength{\tabcolsep}{3pt}
\resizebox{\columnwidth}{!}{%
\begin{tabular}{l ccc ccc}
\toprule
& \multicolumn{3}{c}{Qwen3-30B-A3B-Thinking} & \multicolumn{3}{c}{Gemini-3.1-Pro} \\
\cmidrule(lr){2-4} \cmidrule(lr){5-7}
Level & DSL & Direct & Undec. & DSL & Direct & Undec. \\
\midrule
Expert       & \best{43.8\%} & 20.9\% & 35.3\% & \best{13.8\%} & 4.8\%  & 81.4\% \\
Intermediate & \best{25.6\%} & 20.4\% & 54.0\% & \best{17.6\%} & 12.1\% & 70.3\% \\
Beginner     & \best{23.0\%} & 22.1\% & 54.9\% & 17.1\%        & \best{17.5\%} & 65.3\% \\
Abstract     & \best{31.1\%} & 30.1\% & 38.9\% & \best{31.1\%} & 24.0\% & 45.0\% \\
\bottomrule
\end{tabular}%
}
\caption{Text2CAD pairwise VLM-judge preferences between DSL and direct CadQuery generation, across abstraction levels and two models (single seed, 885 examples/level/model). Percentages are over pairs for which a judgment was reached.}
\label{tab:text2cad}
\end{table}

Table~\ref{tab:text2cad} shows that the DSL is preferred in seven of eight model--level comparisons, the exception being Gemini-3.1-Pro at the beginner level, where the two are effectively tied (17.1\% vs.\ 17.5\%). Some of this gain is a matter of reliability: the DSL lowers the execution-failure rate for both models, most sharply at the expert level, from 51.3\% to 35.1\% for Qwen3-30B-A3B-Thinking and from 21.7\% to 13.4\% for Gemini-3.1-Pro.

\section{Can LLMs Induce DSLs Themselves?}
\label{sec:dsl_induction_results}
\begin{table*}[t]
\centering
\small
\setlength{\tabcolsep}{4pt}
{%
\renewcommand{\arraystretch}{0.92}
\setlength{\aboverulesep}{1pt}
\setlength{\belowrulesep}{1pt}
\resizebox{\textwidth}{!}{%
\begin{tabular}{l c c c c c c c c}
\toprule
& \multicolumn{4}{c}{Proprietary Models}
& \multicolumn{4}{c}{Open-Source Models} \\
\cmidrule(lr){2-5}
\cmidrule(lr){6-9}
& \multicolumn{2}{c}{GPT-5.1 (High)}
& \multicolumn{2}{c}{Gemini 3.1 Pro (High)}
& \multicolumn{2}{c}{Qwen3 30B Thinking}
& \multicolumn{2}{c}{Gemma3 27B Instruct} \\
\cmidrule(lr){2-3}
\cmidrule(lr){4-5}
\cmidrule(lr){6-7}
\cmidrule(lr){8-9}
& F1 & EM & F1 & EM & F1 & EM & F1 & EM \\
\midrule
Minimal DSL
& 0.740{\scriptsize$\pm$0.008} & 0.282{\scriptsize$\pm$0.034}
& 0.839{\scriptsize$\pm$0.033} & 0.484{\scriptsize$\pm$0.028}
& 0.542{\scriptsize$\pm$0.003} & 0.145{\scriptsize$\pm$0.016}
& 0.371{\scriptsize$\pm$0.019} & 0.097{\scriptsize$\pm$0.000} \\
All-at-once
& \runner{0.757{\scriptsize$\pm$0.082}} & 0.314{\scriptsize$\pm$0.103}
& 0.776{\scriptsize$\pm$0.069} & 0.321{\scriptsize$\pm$0.099}
& \runner{0.565{\scriptsize$\pm$0.019}} & \runner{0.152{\scriptsize$\pm$0.018}}
& 0.416{\scriptsize$\pm$0.055} & \runner{0.129{\scriptsize$\pm$0.023}} \\
Batches
& 0.745{\scriptsize$\pm$0.088} & \runner{0.339{\scriptsize$\pm$0.160}}
& \runner{0.828{\scriptsize$\pm$0.032}} & \runner{0.518{\scriptsize$\pm$0.023}}
& 0.480{\scriptsize$\pm$0.041} & 0.100{\scriptsize$\pm$0.034}
& 0.389{\scriptsize$\pm$0.060} & 0.105{\scriptsize$\pm$0.034} \\
Semantic Batches
& 0.704{\scriptsize$\pm$0.007} & 0.225{\scriptsize$\pm$0.022}
& 0.788{\scriptsize$\pm$0.077} & 0.396{\scriptsize$\pm$0.129}
& 0.531{\scriptsize$\pm$0.025} & 0.149{\scriptsize$\pm$0.019}
& \runner{0.441{\scriptsize$\pm$0.020}} & 0.121{\scriptsize$\pm$0.011} \\
Human-Curated DSL
& \best{0.831{\scriptsize$\pm$0.010}} & \best{0.441{\scriptsize$\pm$0.025}}
& \best{0.861{\scriptsize$\pm$0.016}} & \best{0.591{\scriptsize$\pm$0.019}}
& \best{0.663{\scriptsize$\pm$0.019}} & \best{0.237{\scriptsize$\pm$0.033}}
& \best{0.512{\scriptsize$\pm$0.007}} & \best{0.140{\scriptsize$\pm$0.007}} \\
\bottomrule
\end{tabular}%
}%
}
\caption{Results for induced DSL variants and the human-curated DSL. We report F1 and board-level exact match (EM) as mean $\pm$ standard deviation across three independently induced libraries; each library is evaluated over three execution runs. Gemini rows use Gemini 3.1 Pro and are not comparable to Table~\ref{tab:main-results}.}

\label{tab:dsl-induction}
\end{table*}

Observing the substantial gains using {\sc HexaGen}, our human-curated DSL, we ask whether LLMs can induce  useful and effective DSLs themselves. 

\paragraph{Experimental Setup} We design an experiment where each evaluated model induces its own DSL. 
Each induction variant is run 3 times per model, producing 3 independently induced libraries. Each library is evaluated over 3 execution runs. For proprietary models, we use only the \texttt{high reasoning} setting. 

\paragraph{Induction} 
As mentioned in Section \ref{sec:experimental_setup}, the induction pipeline has three stages: \emph{abstraction discovery}, \emph{method implementation}, and \emph{documentation authoring}, which we detail here in turn. 

In the {\em discovery} stage, we vary how the instruction corpus is presented to the model: \emph{(1) flat}, which seeks and proposes abstractions for all instructions in a single prompt; \emph{(2) arbitrary batching}, which partitions the corpus into fixed-size batches, analyzes each independently, and then deduplicates and merges the resulting proposals; and \emph{(3) semantic batching}, which replaces arbitrary batches with sets of {\em similar} instructions, obtained by $k$-means clustering over BGE-M3 instruction embeddings \citep{bgem3}, targeting batches of   100 instructions for both batched variants.
Grouping related examples before prompting resembles the clustered-sampling variant of \citet[Appendix C.2]{lilo_2024}.

In the {\em implementation stage}, given the proposed method signatures, the model receives the  foundation-library source and doc and generates complete implementations, including docstrings and executable examples. Methods that fail to compile or fail their doctest examples are removed. Validation and deduplication are standard in LLM tool-creation pipelines \citep{craft_2024}.

In the {\em documentation stage}, the model produces the final documentation, which serves as the only interface provided for inference. It is written for LLMs rather than human readability, favoring clarity and unambiguous information.
This  follows LILO's \emph{AutoDoc} procedure \citep{lilo_2024}.

\paragraph{Results.}
Table~\ref{tab:dsl-induction} shows that the human-curated \textsc{HexaGen} achieves the highest F1 and exact match in all four model settings, outperforming the best induced variant by 3--10 F1 points. Still, some induced DSLs perform strongly in individual runs: all-at-once induction with GPT-5.1 (High) reaches 0.816 F1 / 0.452 EM. The main limitation is thus consistency rather than capability, as the same induction settings that produce strong libraries also produce runs falling below the minimal baseline. For Gemini 3.1 Pro the pattern is stronger: no induced variant beats the minimal API they extend.\footnote{Our induced libraries fare better than the negative result of \citet{lilo_2024}, though the comparison is loose: we induce Python functions over a foundation API, a materially easier target than abstractions in a novel formalism (\S\ref{sec:related}).}

Beyond execution quality, the strategies differ in how many methods they induce: {\em batched} discovery yields substantially more methods than {\em flat} analysis, though library size does not correlate with downstream performance (Appendix~\ref{app:method-counts}).
\amit{Added Analysis}

Finally, we manually inspect two induced libraries to learn about the type of errors they produce. This shows how a single foundational error in the library's design can make an induced library perform worse than no library at all. Both come from the same setup, GPT-5.1 (High) with the \emph{flat} induction variant, yet Run~A reaches 0.816 board-F1 while Run~B reaches 0.700, below both non-DSL baselines for this model (0.765 Tiles, 0.757 Unrestricted Python). They cover about the same abstractions and differ in one fundamental choice: Run~B falsely treats the board as a rectangular lattice. Seven of its ten methods rest on that assumption, and all seven run without failing while producing wrong tiles. Run~A gets the geometry right, and its remaining distance to \textsc{HexaGen} is of a different kind: it has no compositional layer, no first-class \texttt{Shape} with set operations, spatial queries, or anchor-based copy (Table~\ref{tab:dsl_capability_coverage}).

\begin{table*}[t]
\centering
\small
\setlength{\tabcolsep}{10pt}
\renewcommand{\arraystretch}{1.2}
\begin{tabular}{@{}lccc@{}}
\toprule
Capability & \textsc{HexaGen} & Run~A & Run~B \\
\midrule
Shape constructors (rays, lines, rings, triangles, fill) & \cmark & \cmark & \xmark \\
Rotation / mirror / translation & \cmark & \cmark & \xmark \\
Hex distance metric & \cmark & \cmark & \absmark \\
Hex adjacency (neighbors) & \cmark & \cmark & \xmark \\
Compositional \texttt{Shape} algebra & \cmark & \absmark & \absmark \\
Spatial / topological queries & \cmark & \absmark & \absmark \\
Tiling / anchor-based copy & \cmark & \absmark & \xmark \\
Board \& color queries & \cmark & \absmark & \absmark \\
\midrule
Board-F1 & $0.831$\,{\scriptsize$\pm 0.010$} & $0.816$ & $0.700$ \\
\bottomrule
\end{tabular}
\caption{Capability coverage and board-F1 for \textsc{HexaGen} and the two GPT-5.1 (High) all-at-once induced DSLs. \cmark\ = supported, \absmark\ = absent, \xmark\ = present but broken.}
\label{tab:dsl_capability_coverage}
\end{table*}

\section{Related Work}
\label{sec:related}
\amit{Moved future work to limiation - it makes sense and it also saves space since Limitations is free}
{\bf Grounded instruction following} maps natural language to executable actions in situated environments, spanning navigation \citep{walk-the-talk, chen_mooney_2011, anderson_2018_vision}, robotic manipulation \citep{tellex_2011, saycan2022arxiv}, and collaborative construction \citep{walk-the-talk, artzi-zettlemoyer-2013-weakly}. Across these settings, performance degrades as instructions become more abstract \citep{ALFRED20, ruis2020gscan, wu-etal-2021-reascan, saycan2022arxiv}. In \textsc{Hexagons} \citep{lachmy-etal-2022-draw}, where instructions are annotated by abstraction level, this trend is measured directly. We take this gap as a starting point and ask whether an executable target language that makes abstractions explicit can help close it.

A growing body of work treats {\bf grounded reasoning as program synthesis:} instead of predicting outputs directly, the model generates executable code. This paradigm has been effective in mathematical reasoning \citep[e.g.,][]{gao_pal_2023, chen2022program}, visual question answering \citep{Gupta2022VisProg, vipergpt}, and embodied control \citep{codeaspolicies2022, singh2022progprompt, saycan2022arxiv}. Code generation is a natural fit for abstract instructions because programming languages already provide mechanisms such as variables, loops, and functions. However, when models generate in a general-purpose language or over low-level action sets, they must still recover task-specific abstractions on their own. Our comparison of direct inference, unrestricted Python, and DSL-based generation shows that making these abstractions explicit and executable is itself a major factor in execution quality.

{\bf Domain-specific abstractions for code generation} have often taken the form of \emph{external} DSLs that restrict the target language, both in classical program synthesis and semantic parsing~\citep{gulwani_2011, berant-etal-2013-semantic} and in recent LLM systems that translate natural language into formal domain-specific targets~\citep{wong2023wordmodelsworldmodels}. \textsc{HexaGen}, by contrast, is an \emph{internal} DSL: the model writes unrestricted Python, while the DSL supplies callable abstractions such as named shapes, spatial queries, and composable transformations. In this respect, it is closer to API-augmented generation \citep{codeaspolicies2022, singh2022progprompt}, though those systems expose fixed low-level action primitives rather than abstractions aligned with the level at which people give instructions.

The question of whether models can {\bf discover reusable abstractions} connects to prior work on library learning and tool induction, which differs from ours in the signal abstractions are found in: recurring code rather than recurring language. In the external DSL setting, DreamCoder \citep{dreamcoder} compresses solved programs, keeping an abstraction when it shortens the corpus by more than it costs to define; LAPS \citep{laps_2021} adds word alignments to that objective, but only to score candidates compression already extracted; and LILO \citep{lilo_2024} adds an LLM searcher and documentation, reporting abstractions proposed by an LLM prompted with programs as an appendix negative result only. On the internal side, LATM~\citep{tool_cai_2024} and CREATOR~\citep{qian-etal-2023-creator} build per-task tools rather than a library, while Voyager~\citep{wang_2023_voyager} admits skills by execution success during open-ended play. We instead induce a library from instructions paired with board states, with no solved programs to compress: what recurs across the instructions is the only signal available, and the abstractions we seek are those a speaker compresses into a phrase --- not the fragments that recur across solutions.

\section{Conclusions}
The core challenge in grounded instruction following is mapping abstract, underspecified language to concrete, executable actions. We show that a DSL is a useful device for making task-relevant abstractions explicit. Concretely, we show  on  \textsc{Hexagons} that generating executable code with \textsc{HexaGen}, a human-curated DSL, substantially outperforms both direct prediction and unrestricted code generation ---  with higher gains at higher abstraction levels. However, current LLMs still do not reliably induce such DSLs with similar efficacy on their own right. Model-authored DSLs in our settings occasionally approach the human-curated baseline but remain far more unstable, with no discovery strategy consistently resolving this instability. Enabling models to automatically construct robust, transparent, task-specific abstraction layers therefore remains an intriguing open challenge. To promote research progress on this matter, we release our code, data, evaluation pipeline, and all DSLs, encouraging the community to advance work on executable abstraction and DSL induction.

\section*{Limitations}

\paragraph{Benchmark coverage.} Our evaluation covers \textsc{Hexagons} and \textsc{Text2CAD}. The two differ in domain and output space, but broader coverage across additional testbeds would strengthen the generality of our claims. The \textsc{Text2CAD} replication is also deliberately lightweight: its DSL comprises eight functions, against the 41 of \textsc{HexaGen}. It therefore establishes that the effect transfers, and does not measure how large that effect can be in a second domain.

\paragraph{Induction without execution feedback.} All three induction strategies share one design: the model authors a DSL from a fixed pool of examples and never runs the result. Run~B suggests this may matter --- its rectangular-lattice assumption breaks seven of ten methods and would surface on the first execution --- though a single case does not establish how much an iterative variant would recover.

\paragraph{Future work.} Two directions follow from these results. The first is multi-turn interaction, which our setting does not consider. The second is DSL induction itself. Our results establish that a good abstraction layer helps and that current models do not reliably produce one, which leaves the induction problem open.


\bibliography{custom}

\appendix

\lstset{
  basicstyle=\ttfamily\scriptsize,
  breaklines=true,
  showspaces=false,
  showtabs=false,
  showstringspaces=false,
  keepspaces=true,
  postbreak={},
  tabsize=2,
  columns=fullflexible,
  frame=single,
}

\section{Visual Feedback}
\label{app:vision}

We test whether giving the model visual access to the board it is constructing improves execution.

\noindent\textbf{Setup.} We use the \emph{step-by-step} protocol (Appendix~\ref{app:full_procedure_benchmark}) rather than our full-procedure formulation, since the latter emits the entire procedure in one turn and leaves no point at which an intermediate board could be shown. After each instruction is executed, we render the resulting board and supply the image alongside the next instruction. We evaluate both execution strategies, \emph{Tiles} and \textsc{HexaGen}, with and without the rendered board, using Gemini~3.1~Pro on the validation set.

\begin{table}[H]
\centering
\footnotesize
\setlength{\tabcolsep}{4pt}
\resizebox{\columnwidth}{!}{%
\begin{tabular}{l c c c c}
\toprule
Setup & EM & B-F1 & A-F1 \\
\midrule
Tiles & 0.590 &  0.864 & 0.819 \\
Tiles + Vision & 0.464 & 0.830 & 0.782 \\
\midrule
\textsc{HexaGen} & \best{0.744} & \best{0.902} & \best{0.867} \\
\textsc{HexaGen} + Vision & 0.727 &  0.896 & 0.860 \\
\bottomrule
\end{tabular}%
}
\caption{Effect of visual feedback under step-by-step prompting (Gemini 3.1 Pro, validation set, 446 instructions). EM is board-level exact match; B-F1 and A-F1 are board- and action-level F1.}
\label{tab:vision-feedback}
\end{table}

\noindent\textbf{Results.} Visual feedback does not help either strategy. For \textsc{HexaGen} the difference is negligible; for direct tile prediction it costs over twelve points of exact match. We report this as a negative preliminary result, and leave the interactive setting --- where the model can act on visual feedback rather than only observe it --- to future work.

\section{Full-Procedure vs.\ Step-by-Step Prompting}
\label{app:full_procedure_benchmark}

To determine the querying setup for our generation framework, we compared our \textit{full-procedure} formulation against the default \textsc{Hexagons} evaluation protocol. The original benchmark uses step-by-step prompting with access to the full conversational history, whereas our framework provides the model with the entire instruction sequence at once.

As shown in Table~\ref{tab:benchmark-comparison}, full-procedure prompting performs comparably to the original benchmark protocol across both tested models, and is even slightly better. We therefore use it throughout. This choice simplifies the generation pipeline, reduces inference overhead, better matches real-life settings where the full task is available upfront, and avoids conflating execution quality with dialogue-history effects.

\begin{table}[H]
\centering
\footnotesize
\setlength{\tabcolsep}{4pt}
\resizebox{1\columnwidth}{!}{%
\begin{tabular}{l c c c c}
\toprule
& \multicolumn{4}{c}{Original Benchmark Comparison} \\
\cmidrule(lr){2-5}
Setup & GPT-5.1 (Low) & Gemma 3 & DeBERTa & T5 \\
\midrule
Step-by-step + Full History
& 0.520{\scriptsize$\pm$0.003} & 0.404{\scriptsize$\pm$0.002} & 0.404 & 0.386 \\
Full-Procedure
& 0.541{\scriptsize$\pm$0.007} & 0.407{\scriptsize$\pm$0.005} & -- & -- \\
\bottomrule
\end{tabular}%
}
\caption{Comparison to the original \textsc{Hexagons} benchmark in F1.}
\label{tab:benchmark-comparison}
\end{table}

\section{Text2CAD Evaluation Details}
\label{app:text2cad}
\noindent\textbf{Data.} Text2CAD~\citep{text2cad} with CadQuery annotations~\citep{text2cadquery}; the 1{,}000-object test split, each object with instructions at four abstraction levels (\emph{expert}, \emph{intermediate}, \emph{beginner}, \emph{abstract}) and a gold CadQuery program. We retain the 885 objects whose gold program executes and evaluate on these at every level. Gold executions are cached and deduplicated. Single seed.

\noindent\textbf{Conditions.} \emph{Direct}: raw CadQuery. \emph{DSL}: an 8-function hand-curated library over CadQuery, built following the design principles of Section~\ref{sec:dsl}. Both conditions are generated by Qwen3-30B-A3B-Thinking and Gemini-3.1-Pro.

\noindent\textbf{Evaluation.} We apply one protocol at all four levels. Renders of the two conditions are shown paired to Qwen3.6-27B as a VLM judge, which selects the better match to the instruction or returns \emph{undecided}; the gold object is never shown and presentation order is randomized. An example on which exactly one condition fails to execute is awarded to the other; where both fail, no comparison is possible and the example is dropped.

\section{Method Counts Across Induction Strategies}
\label{app:method-counts}
\begin{figure*}[t]
  \centering
  \includegraphics[width=0.85\textwidth]{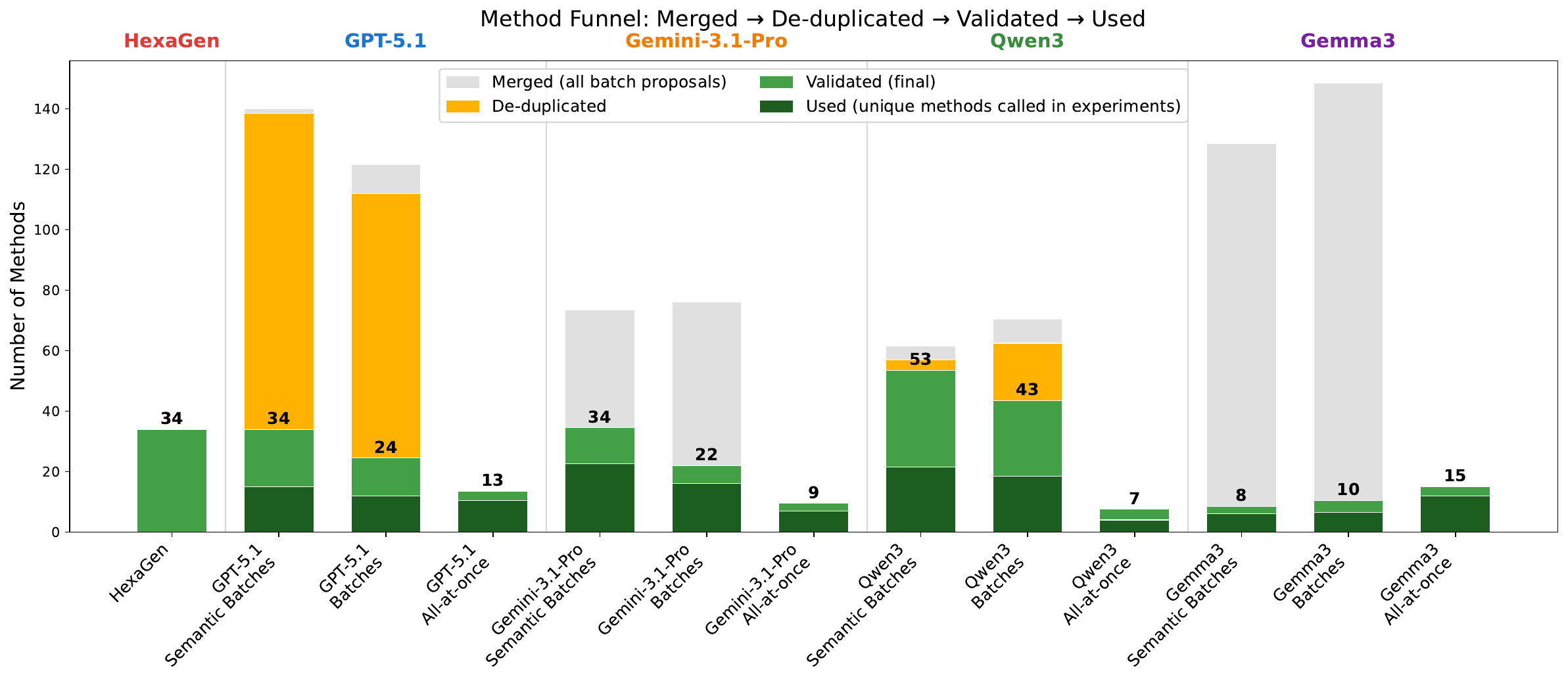}
  \caption{For each model and induction variant, we report how many methods remain after each pipeline stage: merge, deduplication, validation, and downstream use.}
  \label{fig:method-funnel}
\end{figure*}

Figure~\ref{fig:method-funnel} tracks how many candidate methods survive each stage of induction --- proposal, deduplication, and validation --- under the three induction strategies. For the three stronger models, both batched strategies produce substantially more validated methods than flat analysis, suggesting that a single pass over the full instruction corpus is too coarse for the model to attend to recurring patterns. Gemma~3 is the least consistent, with method counts varying greatly across strategies. A larger library is not in itself better, however: many candidates are removed during deduplication or validation, and only a subset is ever called downstream. Method count is therefore best read as a diagnostic of the induction process rather than as a quality measure, with the execution results in Table~\ref{tab:dsl-induction} providing the relevant comparison.

\section{HexaGen API}
\label{app:hexagen_api}

Table~\ref{tab:hexagen_methods} lists the public classes, methods, properties, and operators available in the \textsc{HexaGen} DSL. This is the full API exposed to the language model; internal mathematical and coordinate helper functions remain private and are not accessible during generation.

\noindent\textbf{Constants.}
The library also exposes the constants \texttt{WIDTH=18}, \texttt{HEIGHT=10}, the color set
\texttt{\{white, black, yellow, green, red, blue, purple, orange\}}, and the six hex-grid directions
\texttt{\{up, down, up\_right, down\_left, up\_left, down\_right\}}.

\begin{table*}[p]
  \centering
  \small
  \setlength{\tabcolsep}{5pt}
  \begin{tabular}{l l p{0.56\linewidth}}
    \toprule
    \textbf{Class} & \textbf{Method / Property} & \textbf{Description} \\
    \midrule

    \textbf{\texttt{Game}} 
    & \texttt{\_\_init\_\_(width, height)} & Initializes a game with board dimensions (defaults to \texttt{WIDTH=18}, \texttt{HEIGHT=10}). \\
    & \texttt{start(width, height)} & Factory method returning a new game instance. \\
    & \texttt{\_\_enter\_\_()}, \texttt{\_\_exit\_\_(...)} & Context-manager interface for writing programs inside a \texttt{Game} block. \\
    & \texttt{record\_step(name)} & Saves a snapshot of the current painted tiles under a step name. \\
    & \texttt{get\_record(names)} & Returns the tiles painted in one or more named steps. \\
    & \texttt{plot(...)} & Visualizes the board state, optionally against a gold reference. \\
    \midrule

    \textbf{\texttt{Shape}}
    & \texttt{\_\_init\_\_(...)} & Constructs a shape from tiles or other shapes. \\
    & \texttt{draw(color)} & Paints all active tiles in the shape with the specified color. \\
    & \texttt{copy\_paste(...)} & Copies the shape along a direction, shift vector, or source--destination mapping. \\
    & \texttt{reflect(...)} & Reflects the shape across a line, coordinate axis, or directional axis. \\
    & \texttt{grid(...)} & Repeats the shape multiple times along a direction. \\
    & \texttt{rotate(center, angle)} & Rotates the shape around a tile by a multiple of 60 degrees. \\
    & \texttt{recolor(color\_map)} & Recolors tiles inside the shape according to a mapping dictionary. \\
    & \texttt{get(criterion)} & Selects tiles satisfying a geometric or color criterion (e.g., \texttt{outside}, \texttt{above}, a direction, or a color). \\
    & \texttt{neighbors(criterion)} & Returns neighboring tiles satisfying a criterion such as \texttt{all}, \texttt{white}, \texttt{inside}, or a direction. \\
    & \texttt{boundary(...)} & Returns the shape's boundary tiles, including \texttt{outer} or \texttt{inner} boundary modes. \\
    & \texttt{edge(direction)} & Returns the outermost boundary segment in a specified direction. \\
    & \texttt{extreme(direction)} & Returns the furthest tile in a specified hex direction. \\
    & \texttt{intersection(other)} & Returns the tiles shared by this and another shape. \\
    & \texttt{overlaps(other)} & Returns whether this shape shares any tiles with another. \\
    & \texttt{is\_empty()} & Returns whether the shape contains no tiles. \\
    & \texttt{center()} & Returns the center-of-mass tile of the shape. \\
    & \texttt{\_\_add\_\_}, \texttt{\_\_sub\_\_}, \texttt{\_\_mul\_\_} & Set-style union, difference, and intersection operators over shapes. \\
    & \texttt{get\_entire\_board()} & Static method returning a shape containing all board tiles. \\
    & \texttt{get\_board\_perimeter()} & Static method returning the perimeter of the board. \\
    & \texttt{get\_color(color)} & Static method returning all active tiles of a given color (or all colors). \\
    & \texttt{get\_column(column)} & Static method returning all tiles in a given column. \\
    & \texttt{polygon(...)} & Static method connecting consecutive vertices with \texttt{Line}s. \\
    \midrule

    \textbf{\texttt{Tile}}
    & \texttt{\_\_init\_\_(row, column)} & Creates a tile at the specified 1-based coordinates. \\
    & \texttt{neighbor(direction)} & Returns the adjacent tile in a hex-grid direction. \\
    & \texttt{on\_board()} & Returns whether the tile lies on the board. \\
    & \texttt{color}, \texttt{row}, \texttt{column} & Properties returning the current color and 1-based coordinates. \\
    & \texttt{offset} & Property returning the \texttt{(row, column)} pair. \\
    \midrule

    \textbf{\texttt{Line}}
    & \texttt{\_\_init\_\_(...)} & Creates a line from a start tile and either an end tile or a direction and length. \\
    & \texttt{parallel(shift, space)} & Returns a parallel line translated by the specified spacing. \\
    \midrule

    \textbf{\texttt{Circle}}
    & \texttt{\_\_init\_\_(center, radius)} & Creates a circular outline around a center tile with a given radius. \\
    \midrule

    \textbf{\texttt{Triangle}}
    & \texttt{\_\_init\_\_(...)} & Creates a directional triangular border shape from a start tile, orientation, and side length. \\
    \bottomrule
  \end{tabular}
  \caption{Public classes, methods, properties, and operators exposed in the \textsc{HexaGen} base library.}
  \label{tab:hexagen_methods}
\end{table*}

\section{Minimal API}
\label{app:minimal_api}

Table~\ref{tab:minimal_methods} lists the public classes, methods, and properties available in the minimal \textsc{HexaGen} API. This is the full API exposed to the language model; internal helper classes and mathematical or coordinate utilities remain private and are not accessible during generation.

\begin{table}[H]
  \centering
  \footnotesize
  \setlength{\tabcolsep}{2.5pt}
  \renewcommand{\arraystretch}{1.05}

  \begin{tabularx}{\columnwidth}{
    @{}
    >{\raggedright\arraybackslash}p{0.11\columnwidth}
    >{\raggedright\arraybackslash}p{0.30\columnwidth}
    >{\raggedright\arraybackslash}X
    @{}
  }
    \toprule
    \textbf{Class} &
    \textbf{Method / Property} &
    \textbf{Description} \\
    \midrule

    \textbf{\texttt{Game}}
    & \texttt{\_\_init\_\_(width, height)}
    & Initializes the board with the given dimensions. \\

    & \texttt{start(width, height)}
    & Creates a new game instance with the given dimensions. \\

    & \texttt{\_\_enter\_\_(), \_\_exit\_\_(...)}
    & Supports context-managed execution inside a \texttt{Game} block. \\

    & \texttt{plot(...)}
    & Visualizes the board state, optionally against a gold reference. \\

    \midrule

    \textbf{\texttt{Tile}}
    & \texttt{\_\_init\_\_(row, column, game)}
    & Creates a tile at the specified coordinates. \\

    & \texttt{draw(color)}
    & Paints the tile with the specified color. \\

    & \texttt{color, row, column}
    & Properties returning the current color and 1-based coordinates. \\

    \bottomrule
  \end{tabularx}

  \caption{Public classes, methods, and properties exposed in the minimal
  \textsc{HexaGen} API.}
  \label{tab:minimal_methods}
\end{table}

\noindent\textbf{Constants.}
The API also exposes the board dimensions \texttt{WIDTH=18} and \texttt{HEIGHT=10}, the color set
\texttt{\{white, black, yellow, green, red, blue, purple, orange\}}, and the six hex-grid directions
\texttt{\{up, down, up\_right, down\_left, up\_left, down\_right\}}.

\section{Standard Experiment Prompts}
\label{app:experiment_prompts}

\noindent\textbf{Shared game-context system prompt.}
\begin{lstlisting}[language=python, numbers=none]
# ==========  HEXAGONS GAME OVERVIEW  ==========

# Board geometry
  - 18 columns x 10 rows, both **1-indexed** (column 1 = left edge; row 1 = top).
  - Address any tile by the tuple **(row, column)**.
  - Every tile starts as **white** and keeps its color until repainted.

# Colors
  - Allowed colors: ['white','black','yellow','green','red','blue','purple','orange'].

# Game flow
  - A human **Instructor** writes natural-language steps.
  - You are the **Executor** who must ground each step on the board.
  - Previously painted tiles stay as they are unless a step changes them.

# Abstraction note
  - Instructions may employ higher-level abstractions (objects, loops, conditions, symmetry); a single sentence can cover many low-level paint actions.
\end{lstlisting}

\noindent\textbf{Code-generation prompt.}

\textbf{System prompt:}
\begin{lstlisting}[language=python, numbers=none]
# ====  CODE-GENERATION  ====

# Role
  - You are an expert **Python** assistant for the Hexagons drawing game.

# Task
  - Each turn you receive an *open* Python code block that ends with a `# TODO`.
  - Complete the code so it satisfies the `# TODO`, then finish with:
        board_state = g.board_state

# Rules
  1. **Write only Python -- no prose -- until the fence closes.**
  2. Do **not** add imports or create a Game; the script already begins with:
         from hexagen import Game, Tile, Shape, Line, Circle, Triangle
         with Game() as g:
  3. Store every new shape or tile in a descriptive variable
     (e.g. `red_flower`, `flower_stack`) so later steps can reuse them.
  4. Write all lines of code using the *same* **4-space indentation** that the `# TODO` line uses; indent the first line as well. This instruction is super important as the code won't run without the proper indentations.
  5. Never overwrite previously defined variables and add **no** extra comments, prints, or plots.
  6. Columns and rows are **1-indexed**.
  7. Use **only** the functions, classes, and parameters explicitly defined in the provided library API specification. Do not invent, guess, or call any methods, arguments, or attributes that are not listed there.

[... Injected HexaGen API Spec ...]
\end{lstlisting}

\textbf{User message template:}
\begin{lstlisting}[language=python, numbers=none]
##### CONTEXT (completed instructions so far)
{HISTORY_BLOCK}

##### CURRENT SCRIPT
{CODE_SO_FAR}
{NEXT_STEP}
\end{lstlisting}

\noindent\textbf{Tile-prediction prompt.}

\textbf{System prompt:}
\begin{lstlisting}[language=python, numbers=none]
# =====  TILE-SELECTION  =====

# Role
  - You are an assistant for a hexagonal drawing game.

# Task
  - Given natural-language instructions, output which tiles should be colored.

# Return format
  - A list of tuples in the form **(row, column, color)**.

# Rules
  1. Rows and columns start at **1**.
  2. Allowed colors: ['white','black','yellow','green','red','blue','purple','orange'].
  3. Reply **only** with the list of tuples -- no extra prose.
\end{lstlisting}

\textbf{User message template:}
\begin{lstlisting}[language=python, numbers=none]
##### CONTEXT (completed instructions so far)
{HISTORY_BLOCK}

##### NEXT INSTRUCTION
{NEXT_STEP}

Provide the tiles as a list of (row,column,color) tuples:
\end{lstlisting}

\noindent\textbf{Unrestricted-Python ablation prompt.}

\textbf{System prompt:}
\begin{lstlisting}[language=python, numbers=none]
# =====  PYTHON-BASED TILE GENERATION  =====

# Role
  - You are an assistant for a hexagonal drawing game.

# Task
  - Given natural-language instructions, write unrestricted Python code to compute which tiles should be colored.
  - Output the tiles as a list of tuples in the form **(row, column, color)**.

# Constraints
  - You may use any Python logic (loops, conditions, functions, list comprehensions, etc.).
  - You are NOT given a drawing DSL or library. Use pure Python to calculate tile positions.
  - The board has 10 rows and 18 columns, both **1-indexed**.

# Return format
  - Write a Python function named `get_tiles()` that returns a list of (row, column, color) tuples.
  - The function must return the result as a list: `return [(row, col, color), ...]`
  - Reply **only** with Python code -- no extra prose before or after.
  - Do not include markdown code blocks.
\end{lstlisting}

\textbf{User message template:}
\begin{lstlisting}[language=python, numbers=none]
##### CONTEXT (completed instructions so far)
{HISTORY_BLOCK}

##### INSTRUCTIONS
{NEXT_STEP}

Write Python code with a `get_tiles()` function that returns the list of (row, column, color) tuples:
\end{lstlisting}

\noindent\textbf{HexaGen API specification.}
\begin{lstlisting}[language=python, numbers=none]
# ====  HEXAGEN API SPEC  ====
# One entity per line; two-space indentation shows containment.
# Extra 'Notes:' lines clarify hidden behaviour.

Import: from hexagen import Game, Tile, Shape, Line, Circle, Triangle

Constant: WIDTH = 18
Constant: HEIGHT = 10
Constant: COLORS = ['white','black','yellow','green','red','blue','purple','orange']

# These are the hexagon board directions, most methods use these directions, some of them also extends the list to more directions.
Constant: DIRECTIONS = ['up','down','up_right','down_left','up_left','down_right']

# ------
Class: Game
  Method: __init__(width:int = WIDTH, height:int = HEIGHT)
  Method: start(width:int = WIDTH, height:int = HEIGHT) -> Game
  Method: __enter__() -> Game
  Method: __exit__(exc_type, exc, tb) -> bool
  Method: record_step(step_name:str)                        # Notes: closes previous step snapshot
  Method: get_record(step_names:str | list[str]) -> Shape   # tiles drawn in named step(s)
  Method: plot(gold_boards:list[int] | list[list[int]] | None = None,
               multiple:bool = False, file_name:str | None = None, show:bool = True) -> Figure

# ------
Class: Shape           # Base class for collections of tiles
  Method: __init__(tiles:list[Tile] | Tile | Shape, from_linds:bool = False,
                   from_hexagons:bool = False, game:Game | None = None)
    Notes: duplicate tiles removed; single-tile => auto-up-casts to Tile.

  # -- Painting helpers (ALL mutate active board **and** return the new Shape) --
  Method: draw(color:str) -> Shape                                       # auto-paints
    Notes: draw() overwrites the tile's previous color. To find where two shapes overlap,
    compute the intersection BEFORE drawing: overlap = shape_a * shape_b (or shape_a.intersection(shape_b)).
  Method: copy_paste(shift_direction:str | None = None,                   # auto-paints
                     spacing:int = 0, reference_shape:Shape | None = None,
                     source:Tile | None = None, destination:Tile | None = None,
                     shift:_Vec | None = None) -> Shape
    Notes: Returns a NEW Shape at the destination. To chain multiple copies,
    call copy_paste on the RETURNED shape. For repeated copies in one direction, use grid() instead.
    Param options: shift_direction in {'left','right'} OR DIRECTIONS OR None
  Method: reflect(axis_line:Line | None = None, column:int | None = None,  # auto-paints
                  axis_direction:str | None = None, tile_on_axis:Tile | None = None) -> Shape
    Param options: axis_direction in {'horizontal','vertical'} OR DIRECTIONS
  Method: grid(shift_direction:str, spacing:int, num_copies:int | None = None) -> Shape  # auto-paints
    shift_direction in {'left','right'} OR DIRECTIONS OR None; num_copies=None => 'fill board'.
  Method: rotate(center_tile:Tile, angle:int) -> Shape                    # auto-paints
    angle must be multiple of 60.
  Method: recolor(color_map:dict[str,str]) -> Shape                       # auto-paints

  # -- Geometric queries -------------------------------------------------
  Method: get(criterion:str) -> Shape  # the method returns a new Shape object that has the criterion geometric relation to the original shape or the tiles with selected colors
    Param options: {'outside','inside','above','below','top','bottom',
                    'corners','endpoints'} OR DIRECTIONS OR COLORS
  Method: neighbors(criterion:str = 'all') -> Shape
    Param options: {'all','right','left','above','below','outside','inside','white'} OR DIRECTIONS
  Method: boundary(criterion:str = 'all') -> Shape        # criterion in {'all','outer','inner'} # The method returns the boundary tiles belonging to the shape itself, without including any tiles outside the shape.
  Method: edge(direction:str) -> Shape                    # direction in {'left','right','top','bottom'} OR DIRECTIONS
  Method: extreme(direction:str) -> Shape                 # direction in DIRECTIONS
  Method: is_empty() -> bool
  Method: overlaps(other:Shape) -> bool
  Method: intersection(other:Shape) -> Shape              # same as shape_a * shape_b
  Method: center() -> Tile                                # center of mass (rounded)

  # -- Set operators --------------------------------------------------
  Operator: shape_a + shape_b -> Shape                    # union
  Operator: shape_a * shape_b -> Shape                    # intersection
  Operator: shape_a - shape_b -> Shape                    # difference

  # -- Utilities (static) ------------------------------------------------
  Static: get_entire_board(game:Game | None = None) -> Shape
  Static: get_board_perimeter(game:Game | None = None) -> Shape
  Static: get_color(color:str | {'all','any'}, game:Game | None = None) -> Shape
  Static: get_column(column:int, game:Game | None = None) -> Shape
  Static: polygon(vertices:Shape | list[Tile] | Tile, *extra_vertices:Tile) -> Shape
    # CRITICAL: polygon() connects consecutive vertices with Line(), so every
    # consecutive vertex pair must be colinear on a hex axis. If vertices are
    # not hex-colinear, draw each edge as a separate Line(start, direction=..., length=...).

# ------
Class: Tile (inherits Shape)
  Method: __init__(row:int, column:int, game:Game | None = None)
    Notes: negative indices wrap from opposite edge.
  Method: neighbor(direction:str) -> Tile                # direction in DIRECTIONS only -- 'left'/'right' are NOT valid here
  Method: on_board() -> bool
  Property: color:str
  Property: column:int
  Property: row:int
  Property: offset:tuple[int,int]                        # (row,col) 1-based

# ------
Class: Line (inherits Shape)                                      # Note creating a line from two tiles is only valid if the tiles lie on a straight hex direction
  Method: __init__(start_tile:Tile, end_tile:Tile | None = None,
                   direction:str | None = None, length:int | None = None,
                   end_tiles:Shape | None = None,
                   include_start_tile:bool = True, include_end_tile:bool = True)
    Param options: direction in DIRECTIONS only ('left'/'right' NOT valid here); if end_tile supplied, direction/length ignored.
    Notes: If direction is given but length is omitted (None), the line extends to the board edge.
    Only specify length when you need a shorter line.
    # CRITICAL: Line(start, end) raises an error if the two tiles are not colinear
    # on a hex axis (i.e. do NOT share the same q, r, or s cube coordinate).
    # Prefer Line(start, direction=..., length=...) for reliable line creation.
  Method: parallel(shift_direction:str, spacing:int) -> Line      # shift_direction in {'left','right'} OR DIRECTIONS OR None
  Method: draw(color:str) -> Line                                # inherited auto-paints

# ------
Class: Circle (inherits Shape)            # The created Circle has a hollow center
  Method: __init__(center_tile:Tile, radius:int = 1)
  Method: draw(color:str) -> Circle        # inherited auto-paints

# ------
Class: Triangle (inherits Shape)           # The created triangle has only sides with the center beign hollow
  Method: __init__(start_tile:Tile, point:str, start_tile_type:str, side_length:int = 2)
    Param options: point in {'left','right'}; start_tile_type in {'side','top','bottom'}.
    # CRITICAL: side_length must be >= 2. For a single-tile annotation, use Tile.draw() instead.
  Method: draw(color:str) -> Triangle      # inherited auto-paints
\end{lstlisting}

\section{Library Generation Pipeline Prompts}
\label{app:pipeline_prompts}

\noindent\textbf{Stage 1: API discovery.}
\begin{lstlisting}[language=python, numbers=none]
You have a minimal base library:

{base_lib_docs}

Analyze these {len(batch)} instructions and design NEW methods to extend the base library:
{io_note}
{format_instructions(batch)}

OUTPUT FORMAT:

Class: ClassName
  Method: method_name(param: type) -> ReturnType
    Description: [1 sentence - what it does]
    Rationale: [What problem this solves + how often this pattern appears in the instructions]
    Necessity: [Why users need this abstraction]


DESIGN CRITERIA:
Each method must:
- Address a complex pattern found in the instructions above
- Be non-trivial - users would struggle to implement correctly
- Involve non-trivial algorithms or domain-specific logic
- Require knowledge of internal data structures or complex rules

NOT:
- Easily composable from existing methods
- Simple wrapper or utility
- Simple property access, collection builders, thin wrappers

Quality > Quantity. Propose only methods meeting all criteria.
\end{lstlisting}

\noindent\textbf{Stage 3: Method implementation.}
\begin{lstlisting}[language=python, numbers=none]
Implement the NEW methods from this API proposal{domain_context}.

API PROPOSAL:
{api_proposal}

BASE LIBRARY (reference):
...

OUTPUT FORMAT - Use this exact structure:

# CLASS: ClassName
def method_name(self, param1: type1, param2: type2) -> ReturnType:
    '''Brief description.

    Args: ...
    Returns: ...
    Examples:
        >>> obj = ClassName(...)
        >>> obj.method_name(arg1, arg2)
        expected_result
    '''
    # Implementation
    pass


REQUIREMENTS:
- Complete implementations with docstrings
- Each method MUST include at least one doctest example in its docstring
- Handle edge cases (None, empty, out-of-bounds)
- Match base library coding style

Provide ONLY the new method implementations.
\end{lstlisting}

\noindent\textbf{Stage 4: Documentation spec generation.}
\begin{lstlisting}[language=python, numbers=none]
Write the API spec section for class `{class_name}`{domain_context}.

CLASS SOURCE CODE:
...

EXAMPLE INSTRUCTIONS this library handles:
{instructions_text}

OUTPUT FORMAT -- produce ONLY a Class section like this:

Class: ClassName                              # optional brief role comment
  # -- Category Name ------------------------ # group related methods
  Method: method_name(param1:type, param2:type) -> ReturnType    # brief inline note
    Param options: param1 in {'val1','val2'} OR SOME_CONSTANT
    Notes: One sentence max. Only when inline comment isn't enough.
    # CRITICAL: warning about non-obvious constraint or common mistake
...

RULES:
1. Document ONLY public methods (no _ prefix). Skip __repr__, __str__, __hash__, __eq__.
2. Be CONCISE: prefer inline `#` comments over multi-line Notes.
3. GROUP methods by function (e.g. 'Drawing/painting', 'Navigation'). Use `# --` headers.
4. Mark methods that modify state with `# auto-paints` or `# mutates` inline.
5. Add `# CRITICAL:` warnings for methods that are easy to misuse.
6. Use compact set notation for param options.
\end{lstlisting}
\end{document}